\documentclass[11pt]{article}

\usepackage[final]{acl}

\usepackage{times}
\usepackage{latexsym}

\usepackage[T1]{fontenc}

\usepackage[utf8]{inputenc}

\usepackage{microtype}

\usepackage{inconsolata}

\usepackage{graphicx}

\usepackage{booktabs}
\usepackage{array}
\usepackage{multirow}
\usepackage{amsmath}
\usepackage{amssymb}
\usepackage{xspace}
\usepackage{xcolor}
\usepackage{url}
\usepackage{fancyvrb}
\usepackage{fvextra}
\usepackage{soul}
\usepackage{adjustbox}

\newif\ifaimarks
\aimarkstrue
\definecolor{aitextcolor}{HTML}{7B1FA2}
\ifaimarks

  \newenvironment{aitext}{\color{aitextcolor}}{}
\else

\fi

\newcommand{\dggecRho}{3.72\xspace}

\newcommand{\dgcodegenRho}{6.21\xspace}

\newcommand{\dgcodegenCapped}{5\xspace}

\newcommand{\dgcodegenRhoWithCapped}{6.4\xspace}

\newcommand{\dgstoryRho}{11.43\xspace}

\newcommand{\dreamextsummCapped}{40\xspace}

\newcommand{\dreammathCapped}{5\xspace}

\newcommand{\dreamwildbenchCapped}{56\xspace}

\newcommand{\dreamstoryRho}{95.47\xspace}

\newcommand{\dreamstoryCapped}{143\xspace}

\newcommand{\dreamstoryRhoWithCapped}{54.7\xspace}

\newcommand{\dgGroupReproducing}{4.0\xspace}
\newcommand{\dgGroupChained}{5.8\xspace}
\newcommand{\dgGroupOpen}{9.8\xspace}

\newcommand{\dgMaxOverMin}{3.1\xspace}

\newcommand{\dreamGroupReproducing}{22.1\xspace}
\newcommand{\dreamGroupChained}{56.2\xspace}
\newcommand{\dreamGroupOpen}{87.0\xspace}

\newcommand{\dreamMaxOverMinNoSpace}{4.3}

\newcommand{\figOneNItems}{3,860\xspace}

\newcommand{\dgRNovel}{+0.72\xspace}

\newcommand{\dgRS}{+0.76\xspace}
\newcommand{\dgPS}{0.011\xspace}

\newcommand{\dgextsummNovel}{2\%\xspace}

\newcommand{\dgmtS}{8.90\xspace}

\newcommand{\dgmtRhoT}{6.0\xspace}

\newcommand{\dgabssummNovel}{83\%\xspace}
\newcommand{\dgabssummSurprise}{9.2\xspace}

\newcommand{\dgabssummRhoT}{9.5\xspace}

\newcommand{\dgcodegenNovel}{85\%\xspace}
\newcommand{\dgcodegenSurprise}{7.9\xspace}

\newcommand{\dgcodegenRhoT}{6.2\xspace}
\newcommand{\dgmathNovel}{83\%\xspace}
\newcommand{\dgmathSurprise}{7.0\xspace}

\newcommand{\dgmathRhoT}{6.2\xspace}

\newcommand{\dgstoryNovel}{99\%\xspace}

\newcommand{\dgstoryS}{8.86\xspace}

\newcommand{\dgstoryRhoT}{11.4\xspace}
\newcommand{\dreamRNovel}{+0.79\xspace}

\newcommand{\dreamRS}{+0.85\xspace}
\newcommand{\dreamPS}{0.002\xspace}

\definecolor{exthink}{HTML}{009E73}
\definecolor{exanswer}{HTML}{0072B2}

\newcommand{\exNThinkCanvases}{40\xspace}
\newcommand{\exNWriteCanvases}{5\xspace}
\newcommand{\exTokThink}{10,339\xspace}
\newcommand{\exTokWrite}{1,241\xspace}
\newcommand{\exRhoThink}{3.81\xspace}
\newcommand{\exRhoWrite}{2.88\xspace}
\newcommand{\exPassThink}{9.8\xspace}
\newcommand{\exPassWrite}{5.8\xspace}
\newcommand{\exNovelThink}{15\%\xspace}

\newcommand{\exNovelPrompt}{87\%\xspace}

\newcommand{\dfNLabProblems}{160\xspace}

\newcommand{\dfNContests}{5\xspace}

\newcommand{\dfEasyAcc}{79\%\xspace}

\newcommand{\dfMiddleAcc}{57\%\xspace}

\newcommand{\dfHardAcc}{44\%\xspace}

\newcommand{\dfCtlLabWrite}{+1.20\xspace}

\newcommand{\dfCtlSameLenWrite}{+0.83\xspace}

\newcommand{\phNTasks}{10\xspace}
\newcommand{\phRTask}{+0.89\xspace}
\newcommand{\phPTask}{0.001\xspace}

\newcommand{\phgecRhoThink}{5.32\xspace}
\newcommand{\phgecRhoWrite}{3.52\xspace}

\newcommand{\phgecNovel}{1\%\xspace}

\newcommand{\phextsummRhoThink}{3.50\xspace}
\newcommand{\phextsummRhoWrite}{2.07\xspace}

\newcommand{\phextsummNovel}{3\%\xspace}

\newcommand{\phclinnoteNovel}{30\%\xspace}

\newcommand{\phcodeeditNovel}{45\%\xspace}

\newcommand{\phwildbenchNovel}{68\%\xspace}

\newcommand{\phstoryNovel}{76\%\xspace}

\newcommand{\mlNRecords}{5,000\xspace}
\newcommand{\mlEnTokens}{28\xspace}

\newcommand{\mlNLang}{25\xspace}
\newcommand{\mlNFert}{20\xspace}
\newcommand{\mlCheapLang}{Spanish\xspace}
\newcommand{\mlDearLang}{Armenian\xspace}
\newcommand{\mlCheapPasses}{5.3\xspace}
\newcommand{\mlDearPasses}{16.9\xspace}
\newcommand{\mlCheapTokens}{36\xspace}
\newcommand{\mlDearTokens}{75\xspace}
\newcommand{\mlCheapRho}{15.7\xspace}
\newcommand{\mlDearRho}{24.7\xspace}

\newcommand{\mlPassRatioNoSpace}{3.2}

\newcommand{\mlTokRatioNoSpace}{2.1}

\newcommand{\mlRFert}{+0.65\xspace}
\newcommand{\mlPFert}{0.002\xspace}
\newcommand{\mlRChrfFert}{-0.81\xspace}

\definecolor{selfhl}{HTML}{FFD8A8}

\newcommand{\cvNProblems}{255\xspace}

\newcommand{\cvNSeeds}{5\xspace}

\newcommand{\cvMajFive}{71.4\%\xspace}
\newcommand{\cvWFive}{73.7\%\xspace}
\newcommand{\cvThinkFive}{71.0\%\xspace}

\newcommand{\cvGainFiveLo}{0.3\xspace}
\newcommand{\cvGainFiveHi}{4.5\xspace}
\newcommand{\cvGainMin}{2.3\xspace}
\newcommand{\cvGainMax}{4.2\xspace}

\title{Diffusion Language Models Change Speed Across Tasks and Languages}

\author{
    Mohsen Fayyaz$^{1}$ ~ Ali Modarressi$^{2,3}$ ~
    \textbf{Daniel Israel$^{1}$}  ~  \textbf{Lucas Bandarkar$^{1}$}  ~  \textbf{Nanyun Peng$^{1}$} \\
    $^1$ University of California, Los Angeles \\ 
    $^2$ CIS, LMU Munich ~ $^3$ Munich Center for Machine Learning \\
    \texttt{mohsenfayyaz@cs.ucla.edu} ~
}

\begin{document}
\maketitle
\begin{abstract}
Diffusion language models can write several tokens in one forward pass, and they decide during generation how many passes to spend. Two outputs of the same token length can therefore vary a lot in cost. To understand the underlying factors of this cost variation, we analyze two diffusion models across 10 tasks. We find that the number of forward passes per token depends heavily on the task. Text copied or lightly edited from the input is cheap, while open-ended writing is costly. In DiffusionGemma, generating a token for story writing requires more than three times as many forward passes as for grammar correction. Two output properties explain much of this gap: how much of it is new relative to the input, and how rare its words are within the task. The same holds within a single response when thinking is involved. The final answer is cheap when it repeats the model's reasoning trace and costly when it adds something new. In multilingual generation, we find that highly tokenized languages also require more passes per token, adding to the cost they already pay with more tokens. Finally, cost tracks correctness: on the same math problem, wrong answers take more passes per token than correct ones, and weighting candidate answers by decoding speed beats majority voting by up to 4.2 points at no extra cost. Taken together, our results show that diffusion’s computational advantage depends on how much of the output is already determined, while its decoding effort offers a practical signal for making inference more efficient and reliable.
\end{abstract}

\begin{figure}[t]
\centering
\includegraphics[width=\columnwidth]{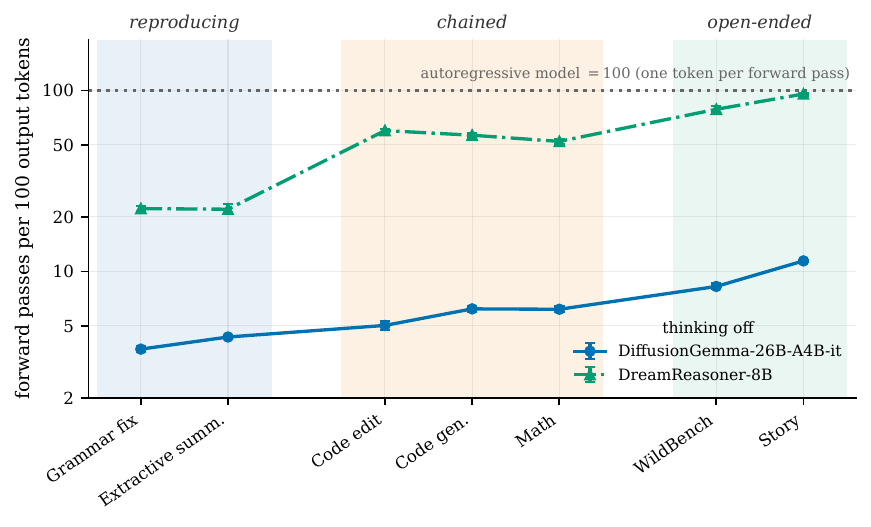}
\caption{Forward passes per 100 output tokens, for seven tasks and two diffusion
models (thinking off). Lower is cheaper. An autoregressive model is always at 100 (dotted line). Error bars are 95\% bootstrap intervals. While architectural differences affect absolute throughput, both models follow the same monotonic cost progression from reproducing to open-ended tasks.
Per-task numbers are in
Table~\ref{tab:forward_passes_by_task_counts}.}
\vspace{-0.1in}
\label{fig:forward_passes_per_100_tokens_by_task}
\end{figure}

\section{Introduction}



Diffusion language models are an increasingly promising alternative to autoregressive language models, offering the potential for substantial inference speedups through parallel token generation \citep{nie2026large, ye2025dream7bdiffusionlarge, diffusiongemma}. While diffusion language models share essentially the same underlying Transformer architecture as autoregressive models \citep{diffusiongemma}, they differ in their decoding process. Rather than generating one token per forward pass, diffusion models iteratively refine groups of tokens, often "committing" a variable number of tokens at each step. For outputs of the same length, the number of forward passes required for decoding can vary substantially, introducing an additional axis for scaling computation \citep{wang2026remasking}. Understanding diffusion decoding effort is therefore important both for characterizing the computational advantages of diffusion models and for achieving maximal efficacy with them in practice.

We systematically study decoding efficiency across tasks and models by measuring the number of forward passes required per generated token. We find that diffusion models are substantially more efficient when generating text constrained by the input or by recurring task patterns, but require considerably more decoding effort on open-ended tasks, such as creative story writing.

Interestingly, we also find that decoding effort is associated with the correctness of mathematical solutions. Incorrect answers require more forward passes per token, even after controlling for problem difficulty and generation length. To better understand how computational effort is distributed within a diffusion model generation, we separately examine reasoning traces and final answers. We find that final answers often require fewer decoding steps, particularly when they reproduce information already established in the reasoning trace. Together, these findings suggest that decoding effort is strongly linked to the remaining uncertainty in the outputs of a diffusion model. 
We additionally investigate multilingual generation and find that languages requiring more tokens also tend to require more decoding passes per token. This compounds the existing tokenization disadvantage \citep{petrov2023, ahia-etal-2023-languages, perry2026tokenizertax} of some languages, producing a \emph{double multilingual tax} on inference efficiency.

Finally, we show in practice that decoding effort is a useful quantity of interest. When sampling multiple candidate solutions for a math problem, weighting each answer's vote by how quickly it was decoded improves accuracy over a plain majority vote \citep{wang2023selfconsistency, brown2024largelanguagemonkeysscaling} by \cvGainMax points with two answers and \cvGainMin points with five, with zero additional overhead.

In summary, our findings characterize when diffusion decoding is most computationally efficient, provide insight into how decoding effort is distributed across tasks and within generations, and motivate inference strategies guided by task structure and denoising signals.


\section{Setup}
\label{sec:setup}

\paragraph{Models.}
We use two diffusion language models:
DiffusionGemma-26B-A4B-it \citep{diffusiongemma}, a 26B mixture-of-experts model trained with uniform diffusion that generates blocks of 256 tokens at a time, and DreamReasoner-8B \citep{dreamreasoner}, an 8B dense model based on masked diffusion that generates blocks of 32 tokens. Both employ semi-autoregressive block diffusion \citep{arriola2025block} and dynamically determine step counts by committing only tokens that meet a target confidence threshold. Adaptive step counts are the basis of decoding effort, which we aim to measure.
Full settings are in Appendix~\ref{app:models_and_settings}.

\paragraph{Cost.}
An autoregressive model computes a forward pass once for every token. A diffusion model
runs the network once per step, and in one step it can output many tokens. For a diffusion output $y$, we count the number of forward passes ($P$) used to generate $y$ and divide by the number of tokens $y$ contains ($T$):
\begin{equation}
\mathrm{cost}(y) = 100 \cdot \frac{P}{T},
\label{eq:cost:forward_passes_per_100_tokens}
\end{equation}
the \emph{forward passes per 100 output tokens}. 
This is the inverse of the tokens per forward pass (TPF) that diffusion decoding typically reports as its speed \citep{fastdllm}. We count only meaningful generated tokens up to the first \texttt{<eos>}, excluding any trailing \texttt{<eos>} or padding tokens, and normalize by generated tokens $T$ rather than block capacity to account for early stopping within blocks and keep cost directly comparable across different block sizes (e.g., 32 vs.\ 256 tokens).
We then compute the cost for each output and
average over the prompts of a task. An autoregressive model always has a cost of 100. A diffusion model that writes 20 tokens per step has cost 5. 

\paragraph{Tasks.}
Diffusion decoding speed is generally reported on a few benchmarks, mostly math and code, as tokens per forward pass or per second \citep{nie2026large, ye2025dream7bdiffusionlarge, bie2025llada20scalingdiffusionlanguage, fastdllm, ben2026accelerated, liu2025dllm}. 
Much less is known about how decoding cost behaves on other kinds of text, and in particular how it depends on the state of the context before the model generates the answer. We select tasks that constrain the answer differently: via the input, the answer's own earlier steps, or very little at all. This gives seven tasks in three groups (Appendix~\ref{app:tasks_and_data}):
\begin{itemize}
\setlength\itemsep{0pt}
\item \textbf{Reproducing}: most of the answer is copied from the input,
  with a few corrections or selected sentences. We use grammar correction
  on W\&I+LOCNESS \citep{bea2019} and extractive summarization on
  GovReport \citep{govreport}.
\item \textbf{Chained}: later parts of the answer build on earlier ones,
  as in code or a step-by-step solution. We use code editing on
  HumanEvalFix \citep{octopack}, code generation on LiveCodeBench
  \citep{livecodebench}, and competition math from MathArena \citep{matharena}.
\item \textbf{Open-ended}: the prompt imposes loose specifications, leaving the output space largely underdetermined. We use user queries from WildBench v2
  \citep{wildbench} and story generation from WritingPrompts \citep{writingprompts}.
\end{itemize}
Starting in Section~\ref{sec:what_makes_an_output_cheap}, we additionally conduct experiments on three
\textbf{rewriting} tasks, where the answer draws on information in the
input but expresses it in new words: German-to-English translation on
WMT23 \citep{wmt23}, clinical notes from doctor--patient conversations
on ACI-BENCH \citep{acibench}, and abstractive chapter summaries on BookSum \citep{booksum}.



\section{Task Dependent Cost}
\label{sec:what_makes_a_token_cheap}
\label{sec:token_cost_depends_on_task}

As previously discussed, a diffusion model does not apply a fixed amount of effort for every task. In Figure~\ref{fig:forward_passes_per_100_tokens_by_task}, we observe that a model with fixed decoding settings utilizes a different number of forward passes per token
depending on the task.

\paragraph{Effort across tasks.}
For DiffusionGemma, the average decoding cost increases across task groups: \dgGroupReproducing forward passes per 100 tokens for reproducing tasks, \dgGroupChained for chained tasks, and \dgGroupOpen for open-ended tasks. At the extremes, grammar correction requires \dggecRho passes, whereas story generation requires \dgstoryRho passes, making story tokens \dgMaxOverMin times as expensive as grammar correction tokens. Between reproducing, chained, and open-ended tasks, a clear hierarchy of decoding effort emerges.

\paragraph{Cross-model agreement on task effort.}
DreamReasoner exhibits a similar monotonic trend across categories, requiring \dreamGroupReproducing, \dreamGroupChained, and \dreamGroupOpen passes per 100 tokens (a \dreamMaxOverMinNoSpace$\times$ spread from cheapest to most expensive). While DreamReasoner incurs higher overall decoding costs due to its smaller block size (32 tokens versus 256), the relative cost ranking remains identical. Two diffusion language models with differing architectures, sizes, and training recipes exhibit agreement on which tasks demand more forward passes.
Figure~\ref{fig:forward_passes_per_100_tokens_by_task} shows that decoding effort depends strongly on the task; tasks that copy or lightly edit the input need far fewer forward passes per token than open-ended writing. This raises the following question: which properties of the output make a token cheap or costly to decode? We study these properties next in more detail.

\definecolor{ttreproducing}{HTML}{0072B2}
\definecolor{ttrewriting}{HTML}{CC79A7}
\definecolor{ttchained}{HTML}{E69F00}
\definecolor{ttopenended}{HTML}{009E73}
\begin{table*}[t]
\centering
\small
\setlength{\tabcolsep}{2.5pt}
\begin{tabular}{l rrrr c rrrr}
\toprule
 & \multicolumn{4}{c}{DiffusionGemma} & & \multicolumn{4}{c}{DreamReasoner} \\
\cmidrule(lr){2-5}\cmidrule(lr){7-10}
task & cost & novelty & surprisal & $\mathrm{NI}$ & & cost & novelty & surprisal & $\mathrm{NI}$ \\
\midrule
\makebox[0.8em][c]{\raisebox{0.35pt}{$\vcenter{\hbox{\textcolor{ttreproducing}{\rule{0pt}{0pt}{\Large$\bullet$}}}}$}}\,Grammar fix & 3.7 & 14\% & 9.1 & 1.30 & & 22.2 & 9\% & 9.1 & 0.82 \\
\makebox[0.8em][c]{\raisebox{0.35pt}{$\vcenter{\hbox{\textcolor{ttreproducing}{\rule{0pt}{0pt}{\Large$\bullet$}}}}$}}\,Extractive summ. & 4.3 & 2\% & 9.7 & 0.19 & & 21.6 & 13\% & 9.7 & 1.25 \\
\makebox[0.8em][c]{\raisebox{0.35pt}{$\vcenter{\hbox{\textcolor{ttchained}{\rule{0pt}{0pt}{\Large$\bullet$}}}}$}}\,Code edit & 5.0 & 52\% & 7.6 & 3.97 & & 59.9 & 81\% & 7.5 & 6.01 \\
\makebox[0.8em][c]{\raisebox{0.8pt}{$\vcenter{\hbox{{\setlength{\fboxsep}{0pt}\setlength{\fboxrule}{0.8pt}\fcolorbox{ttrewriting}{white}{\rule{0pt}{4.2pt}\hspace{4.2pt}}}}}$}}\,Translation & 6.0 & 95\% & 9.3 & 8.90 & & 80.5 & 92\% & 9.5 & 8.68 \\
\makebox[0.8em][c]{\raisebox{0.8pt}{$\vcenter{\hbox{{\setlength{\fboxsep}{0pt}\setlength{\fboxrule}{0.8pt}\fcolorbox{ttrewriting}{white}{\rule{0pt}{4.2pt}\hspace{4.2pt}}}}}$}}\,Clinical note & 6.1 & 77\% & 8.7 & 6.69 & & 83.8 & 89\% & 7.6 & 6.77 \\
\makebox[0.8em][c]{\raisebox{0.35pt}{$\vcenter{\hbox{\textcolor{ttchained}{\rule{0pt}{0pt}{\Large$\bullet$}}}}$}}\,Math & 6.2 & 83\% & 7.0 & 5.82 & & 52.2 & 91\% & 7.4 & 6.77 \\
\makebox[0.8em][c]{\raisebox{0.35pt}{$\vcenter{\hbox{\textcolor{ttchained}{\rule{0pt}{0pt}{\Large$\bullet$}}}}$}}\,Code gen. & 6.2 & 85\% & 7.9 & 6.64 & & 56.5 & 90\% & 7.5 & 6.78 \\
\makebox[0.8em][c]{\raisebox{0.35pt}{$\vcenter{\hbox{\textcolor{ttopenended}{\rule{0pt}{0pt}{\Large$\bullet$}}}}$}}\,WildBench & 8.3 & 90\% & 9.7 & 8.72 & & 77.6 & 89\% & 9.1 & 7.98 \\
\makebox[0.8em][c]{\raisebox{0.8pt}{$\vcenter{\hbox{{\setlength{\fboxsep}{0pt}\setlength{\fboxrule}{0.8pt}\fcolorbox{ttrewriting}{white}{\rule{0pt}{4.2pt}\hspace{4.2pt}}}}}$}}\,Abstractive summ. & 9.5 & 83\% & 9.2 & 7.60 & & 101.0 & 75\% & 8.9 & 6.70 \\
\makebox[0.8em][c]{\raisebox{0.35pt}{$\vcenter{\hbox{\textcolor{ttopenended}{\rule{0pt}{0pt}{\Large$\bullet$}}}}$}}\,Story & 11.4 & 99\% & 9.0 & 8.86 & & 95.5 & 95\% & 8.8 & 8.35 \\
\midrule
Pearson $r$ & & $+0.72$ & $+0.18$ & $\mathbf{+0.76}$ &  & & $+0.79$ & $-0.09$ & $\mathbf{+0.85}$ \\
\quad $p$-value & & \underline{0.020} & 0.623 & \underline{0.011} &  & & \underline{0.007} & 0.800 & \underline{0.002} \\
\bottomrule
\end{tabular}
\caption{Decoding cost (Eq.~\ref{eq:cost:forward_passes_per_100_tokens}), novelty (Eq.~\ref{eq:novelty}), surprisal (Eq.~\ref{eq:surprise}) and new information (NI, Eq.~\ref{eq:new_information}) per task, for DiffusionGemma and DreamReasoner. Task type: \makebox[0.8em][c]{\raisebox{0.35pt}{$\vcenter{\hbox{\textcolor{ttreproducing}{\rule{0pt}{0pt}{\Large$\bullet$}}}}$}}\,reproducing, \makebox[0.8em][c]{\raisebox{0.8pt}{$\vcenter{\hbox{{\setlength{\fboxsep}{0pt}\setlength{\fboxrule}{0.8pt}\fcolorbox{ttrewriting}{white}{\rule{0pt}{4.2pt}\hspace{4.2pt}}}}}$}}\,rewriting, \makebox[0.8em][c]{\raisebox{0.35pt}{$\vcenter{\hbox{\textcolor{ttchained}{\rule{0pt}{0pt}{\Large$\bullet$}}}}$}}\,chained, \makebox[0.8em][c]{\raisebox{0.35pt}{$\vcenter{\hbox{\textcolor{ttopenended}{\rule{0pt}{0pt}{\Large$\bullet$}}}}$}}\,open-ended. 
\textbf{For both models, cost correlates most strongly with NI.}
Rows are sorted by DiffusionGemma’s cost. All measures are computed on each model’s own outputs, excluding items that hit the token limit. The bottom rows give each measure’s Pearson correlation with cost across the ten tasks, within each model (strongest in \textbf{bold}), and its two-sided $p$-value (underlined when below 0.05; uncorrected, and indicative only with ten tasks).
}
\label{tab:novelty_surprise_copy_by_task}
\end{table*}

\subsection{What Makes an Output Cheap to Decode}
\label{sec:what_makes_an_output_cheap}

To understand why certain outputs require fewer decoding steps, we hypothesize two properties of the output text: (1) how much of it is new
relative to the input (\emph{novelty}), and (2) how rare its words are
within the task (\emph{surprisal}). Both are computed per output and averaged within each of the ten tasks (Table~\ref{tab:novelty_surprise_copy_by_task}). Unlike cost, which depends on how a model tokenizes, these concepts do not depend on the tokenizer. We split
the input $x$ and the output $y = (y_1, \ldots, y_{|y|})$ into words rather than model tokens.

\paragraph{Dimension 1: Novelty.}
We segment $y$ into contiguous runs copied directly from $x$ (\emph{matched fragments}) and isolated novel words (\emph{unmatched words}). Processing left to right, each fragment is greedily selected as the longest match in $x$ \citep{grusky2018}. We measure output novelty as the number of resulting segments per output word:
\begin{equation}
    \mathrm{novelty}(y) = \frac{\#^{\text{ matched}}_{\text{ fragments}} + \#^{\text{ unmatched}}_{\text{ words}}}{|y|}.
\label{eq:novelty}
\end{equation}

Novelty approaches 0 when the output consists of a few long copied contiguous fragments, and reaches 1 when the output is generated entirely from scratch (one segment per word). For example, a 10-word output consisting of two matched fragments of 4 and 3 words plus 3 unmatched words contains 5 segments in total, yielding $\text{novelty} = 5/10 = 0.5$.
Novelty is closely related to \emph{copy length} in summarization benchmarks, which measures the average length of copied source fragments \citep{chen2020cdevalsumm,ma2024prom}. However, while copy length averages strictly over matched fragments, novelty additionally treats every uncopied word as an unmatched segment of length 1. For DiffusionGemma, novelty is \dgextsummNovel for extractive summarization compared to \dgstoryNovel for story generation. On its own, novelty correlates strongly with decoding cost ($r = \dgRNovel$ for DiffusionGemma, $r = \dreamRNovel$ for DreamReasoner; Table~\ref{tab:novelty_surprise_copy_by_task}).

\paragraph{Dimension 2: Surprisal.}
A word may be novel with respect to a specific input yet highly predictable within the broader task. For example, code frequently reuses keywords like \texttt{def} and \texttt{return}, boilerplate syntax, and common identifier names; mathematical reasoning repeatedly employs symbols such as \texttt{\textbackslash frac} and \texttt{=}. To quantify this task-level regularity, we aggregate each model's generations for a given task $t$ and compute empirical word frequencies. Letting $c_t(w)$ denote the count of word $w$, $N_t$ the total word count, and $V_t$ the vocabulary size of unique words, we define the add-one smoothed word distribution:
\begin{equation}
P_t(w) = \frac{c_t(w)+1}{N_t+V_t}.
\label{eq:task_word_distribution}
\end{equation}
Following information theory, the surprisal of a word is defined as its self-information $-\log_2 P_t(w)$ \citep{shannon1948, hale2001,levy2008}. A word that makes up half of all words in a task’s outputs costs 1 bit ($-\log_2 0.5 = 1$); a word seen once in a million costs about 20 bits. The surprisal of
an output is the average over its words, so it measures word rarity within the task:
\begin{equation}
\mathrm{surprisal}(y) = -\frac{1}{|y|} \sum_{j=1}^{|y|} \log_2 P_t(y_j).
\label{eq:surprise}
\end{equation}

Across tasks, novelty and surprisal capture largely separate properties. For example, the clearest contrast is math and code generation against abstractive summarization. For DiffusionGemma, all three are mostly
new text (novelty \dgmathNovel, \dgcodegenNovel and \dgabssummNovel), but for math and code, the words are much less surprising (\dgmathSurprise and \dgcodegenSurprise bits, vs.
\dgabssummSurprise), and generation costs much less (\dgmathRhoT and \dgcodegenRhoT forward passes
per 100 tokens, vs. \dgabssummRhoT). 

\begin{figure}[t]
\centering
\includegraphics[width=0.92\columnwidth]{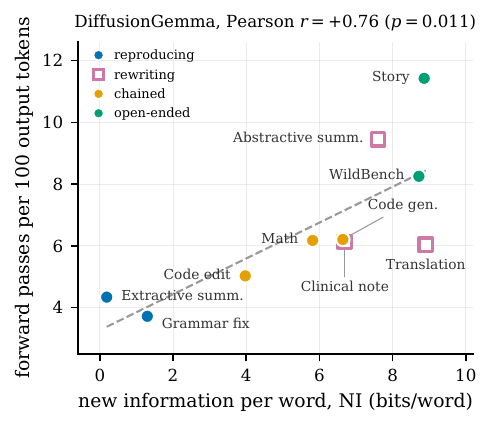}
\caption{Decoding cost against new information (NI) for DiffusionGemma. Each point is one task (values in Table~\ref{tab:novelty_surprise_copy_by_task}); the dashed line shows a least-squares fit.
\textbf{Tasks that produce more new information cost more to decode.}}
\label{fig:cost_vs_new_information_per_word}
\end{figure}

\paragraph{New information (NI).}
Because novelty and surprisal capture separate properties, we combine them into a single measure, the bits of new information per output word:
\begin{equation}
\mathrm{NI}(y) = \mathrm{novelty}(y) \times \mathrm{surprisal}(y)
\label{eq:new_information}
\end{equation}
Copying long sequences verbatim lowers $\mathrm{NI}$ through novelty, and using words that are common in the task lowers it through surprisal. 
Weighting words by rarity follows the NIST machine
translation metric \citep{doddington2002}, which scores $n$-grams by their rarity in the reference translations.

We find that this new metric is justified, as it demonstrates strong predictive power. Cost rises with $\mathrm{NI}$ across tasks (Figure~\ref{fig:cost_vs_new_information_per_word}), and $\mathrm{NI}$ tracks it better than either dimension independently (Table~\ref{tab:novelty_surprise_copy_by_task}), with $r$ rising from \dgRNovel for novelty to \dgRS for $\mathrm{NI}$ ($p=\dgPS$) for DiffusionGemma, and from \dreamRNovel to \dreamRS ($p=\dreamPS$) for DreamReasoner.

\paragraph{Where it does not fit.}
As shown in Figure~\ref{fig:cost_vs_new_information_per_word}, three tasks deviate from the fit. Translation falls below the line:
its $\mathrm{NI}$ matches that of story generation (\dgmtS vs. \dgstoryS for DiffusionGemma), yet it costs \dgmtRhoT vs. \dgstoryRhoT.
Novelty is designed to capture surface copying, so it misses copying across languages: translation reuses the input’s content but not its words.
Story generation and abstractive summarization show the reverse:
their unpredictability lies largely at the discourse level rather than in their words (e.g., narrative surprise in stories), which word-level surprisal cannot capture and which remains hard to measure without human judgment \citep{bissell-etal-2025-theoretical}.



\begin{figure}[t]
\centering
\includegraphics[width=\columnwidth]{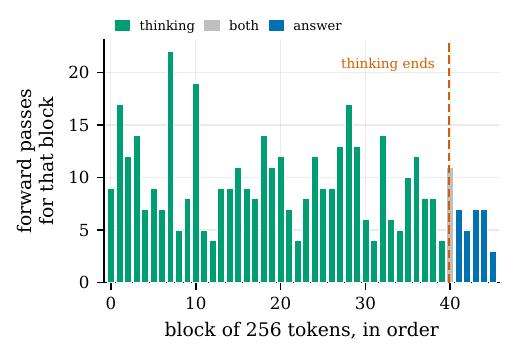}
\caption{One generation with thinking on: forward passes spent on each block of 256
tokens. DiffusionGemma on AIME 2025 problem 2 (answered correctly). Green blocks are
thinking, blue blocks the final answer, and the grey block holds the point where the thinking
ends. Thinking blocks take \textcolor{exthink}{\textbf{\exPassThink}} passes on average, answer blocks \textcolor{exanswer}{\textbf{\exPassWrite}}.}
\vspace{-0.1in}
\label{fig:thinking_example_passes_per_block}
\end{figure}

\section{Analysis: Copying from One's Own Thinking}
\label{sec:thinking_vs_writing}
Reasoning models, which think before they answer, are now common
among autoregressive language models \citep{openai2026openaio1card, deepseekr12025} and have begun to appear among diffusion language
models as well \citep{dreamreasoner, diffusiongemma}. While Section~\ref{sec:what_makes_a_token_cheap} turned
thinking mode off, turning it on provides a new view and interpretation of the previous findings. Once a model has thought, its final
answer can copy from its own thinking, much as a reproducing task
copies from its input. We examine this in a case study on
DiffusionGemma.
It decodes its thinking and its final answer in the same
way, in blocks of 256 tokens, so the cost of each can be measured directly.


\begin{figure}[t]
\centering
\includegraphics[width=\columnwidth]{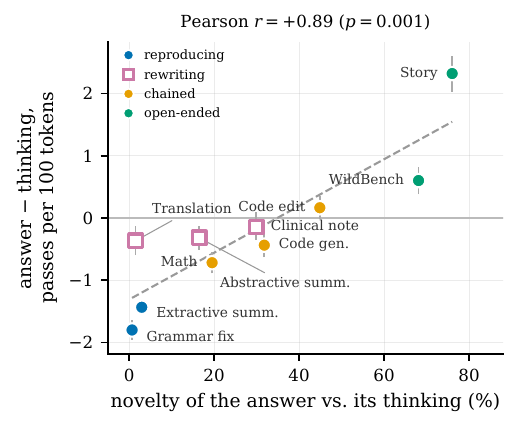}
\caption{The answer is cheap when it copies the thinking. Each point is one task. We plot how novel the answer is relative to the model's own thinking (Eq.~\ref{eq:novelty}) against how much more an answer token costs than a thinking token. Negative values mean answer tokens are cheaper. Error bars are 95\% bootstrap intervals.
All runs use DiffusionGemma with thinking on and its default released settings. 
}
\vspace{-0.1in}
\label{fig:answer_vs_thinking_cost_gap_by_novelty}
\end{figure}

\paragraph{An example.}
Figure~\ref{fig:thinking_example_passes_per_block} follows one competition math problem. The model thinks for
\exTokThink tokens (\exNThinkCanvases blocks, plus one block shared with the answer) and then writes a \exTokWrite-token answer
(\exNWriteCanvases more blocks). Thinking blocks are expensive and uneven, taking from 4 to 22 forward passes (cost \exRhoThink), while final-answer blocks are cheap and even (cost \exRhoWrite). 
In this example, the thinking process has contributed significantly more decoding effort than the final answer. The final answer largely rewords parts of the thinking, reflected in its novelty of \exNovelPrompt against the question but only \exNovelThink against the thinking.

\begin{figure*}[t]
\centering
\includegraphics[width=\textwidth]{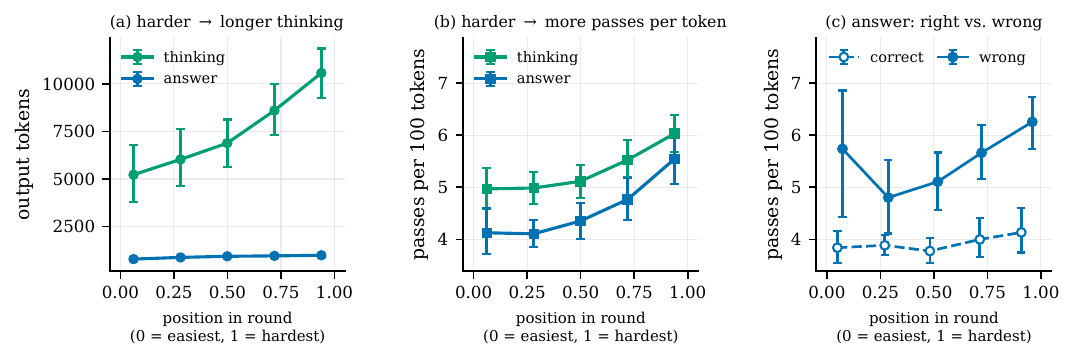}
\caption{\textbf{Harder problems cost more, and wrong answers cost the most.}
We run DiffusionGemma with thinking on over \dfNLabProblems problems from \dfNContests math contests,
five seeds each, and place each problem by its position in the contest round, which runs from easy
to hard. (a) On harder problems the model thinks much longer, while its answer stays about the same
length. (b) Tokens also become more expensive toward the hard end, in both the thinking and the
answer. (c) Looking at the answer alone, wrong answers cost more per token than correct ones at
every level of difficulty. Correct answers stay equally cheap as problems get harder, so the rise
in (b) comes mostly from wrong answers. Error bars are 95\% bootstrap intervals over problems.
}
\vspace{-0.1in}
\label{fig:math_cost_by_difficulty_and_correctness}
\end{figure*}

\subsection{Across Tasks}
\label{sec:across_tasks}
The example suggests that, once the thinking is done, the final answer behaves like a reproducing task with the thinking as its input.
We test this hypothesis on the ten tasks of Table~\ref{tab:novelty_surprise_copy_by_task}, now with thinking on, measuring the final answer’s novelty against the thinking rather than the question.
\emph{Where the final answer copies most, it is clearly cheaper than the thinking}. In grammar correction and extractive summarization, the final answers have a novelty of only \phgecNovel and \phextsummNovel, but it costs \phgecRhoWrite against \phgecRhoThink and \phextsummRhoWrite against \phextsummRhoThink. Translation, abstractive summarization, math, and code generation show the same pattern, albeit by a smaller margin.

More generally, we observe that 
\emph{the less novel the final answer is relative to the thinking, the cheaper it is compared with the thinking} ($r=\phRTask$, $p=\phPTask$;
Figure~\ref{fig:answer_vs_thinking_cost_gap_by_novelty}). 
The same holds within each task, across its generations, where the Pearson correlation between novelty and the gap is positive in all \phNTasks tasks (+0.06 to +0.90) and significant in 7. Thus, it is not only that some tasks have cheaper final answers. Within a task, answers that copy more from their own thinking tend to be cheaper relative to that thinking.

\emph{As novelty rises, the gap closes and then reverses.} For code edit and clinical notes (novelty \phcodeeditNovel and \phclinnoteNovel), the two parts cost about the same. On WildBench and story writing, the final answer is largely new relative to the thinking (novelty \phwildbenchNovel and \phstoryNovel), and it costs more than the thinking (+0.60 and +2.32). The thinking can plan the content, but the sentences are still written for the first time.



\subsection{Across Difficulty}
\label{sec:harder_problems_and_wrong_answers}

A final answer can also remain unsettled when the thinking fails to solve the problem. To vary how often this happens, we use competition-math problems spanning a range of difficulties, drawn from AIME I and II, HMMT, and CMIMC \citep{matharena}. We take a problem’s position within its contest round as a rough difficulty proxy.\footnote{This choice is motivated by competition difficulty-rating conventions: the AoPS competition-rating scheme assigns progressively higher difficulty bands to later AIME problems and to the later half of HMMT February individual rounds, and HARP uses this scheme to assign competition-math difficulty levels \cite{aops_competition_ratings,yue2024harp}. 
Position is a coarse heuristic, not a strict ordering or a calibrated difficulty measure across contests.}
We split each contest round into thirds, labeled easy, middle, and hard, and run each problem with five seeds. Consistent with this proxy, accuracy decreases from \dfEasyAcc on the easy third to \dfMiddleAcc on the middle third and \dfHardAcc on the hard third.

On harder problems, the model thinks much longer. Similar to autoregressive reasoning models, the median thinking grows from 4,416 to 9,710 tokens from the easy to the hard third, while the final answer grows only slightly (Figure \ref{fig:math_cost_by_difficulty_and_correctness}a). Unlike in an autoregressive model, however, where every token costs one forward pass, tokens of both parts also become more expensive (Figure \ref{fig:math_cost_by_difficulty_and_correctness}b): the cost of the thinking rises from 5.00 to 5.89, and that of the final answer from 4.15 to 5.21. The final answer stays cheaper than the thinking at every difficulty, as in Section \ref{sec:across_tasks}, but it rises more steeply, so the gap narrows toward the hard end.

Figure \ref{fig:math_cost_by_difficulty_and_correctness}c shows that this rise is driven primarily by wrong answers. Correct final answers stay equally cheap at every difficulty (3.88 on the easy third, 4.02 on the hard third), while wrong ones cost more and more (5.16 to 6.13). Incorrect final answers also become more prevalent on hard problems (137 of 244 runs on the hard third, against 62 of 297 on the easy third).

Problem difficulty increases the computational cost of generating a final answer, but this effect is driven entirely by incorrect answers. Correct answers remain computationally cheap across all difficulty levels. This disparity is not explained by output length. Controlling for contest, problem position, and token counts across both stages, incorrect answers require \dfCtlLabWrite\ additional forward passes per 100 tokens. Even when holding the problem and output length fixed across different random seeds, incorrect runs still require \dfCtlSameLenWrite\ more passes. 
A qualitative look, as shown in Figure~\ref{fig:self_correction_example_right_vs_wrong}, suggests that when the thinking solves the problem, the final answer retraces the solution. However, when it does not, the model seems to keep working while it outputs the final answer, and wrong final answers stop to correct themselves (“wait”, “let me re-check”) about five times as often as right ones.

\begin{figure}[t]
\centering
\includegraphics[width=\columnwidth]{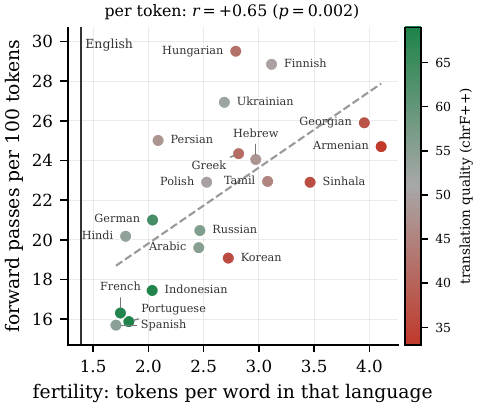}
\caption{Languages that the tokenizer splits into more pieces also cost more per token.
Each point is one target language, translated from the same 200 English FLORES sentences by
DiffusionGemma. We show the \mlNFert languages that separate words with spaces, where tokens per
word is well defined; color indicates translation quality. Results for all \mlNLang
languages are in Fig.~\ref{fig:multilingual_tokens_vs_passes_by_language}.}
\vspace{-0.1in}
\label{fig:multilingual_cost_per_token_vs_fertility}
\end{figure}

\section{Multilingual Effort}
\label{sec:multilingual_tax}

It is well known that the same text costs more tokens in some languages than in others, because
tokenizers split their words into more pieces \citep{petrov2023, ahia-etal-2023-languages, perry2026tokenizertax}. Because autoregressive models always cost one forward pass per token, the additional cost incurred by different languages could be attributed entirely to the tokenizer. In contrast, a diffusion model can also spend a different number of forward passes on each token, so a relevant question is whether forward passes per token
can differ across languages.

We translate the same 200 English FLORES \citep{nllb2022} sentences into \mlNLang languages with DiffusionGemma with thinking off and default settings. The total cost of a sentence varies substantially across languages (Figure~\ref{fig:multilingual_tokens_vs_passes_by_language}). For example, \mlDearLang needs \mlPassRatioNoSpace$\times$
as many forward passes as \mlCheapLang (\mlDearPasses against \mlCheapPasses per sentence). The difference can partly be attributed to token count (\mlDearTokens against \mlCheapTokens tokens, \mlTokRatioNoSpace$\times$), but  
passes per token also play a significant role (\mlDearRho against \mlCheapRho per 100 tokens). 
Languages whose words are split into more tokens also cost more per token
(Figure~\ref{fig:multilingual_cost_per_token_vs_fertility}; $r=\mlRFert$, $p=\mlPFert$, over the \mlNFert languages written with spaces between words). These languages need more tokens, and each token needs more passes. Languages that are split more are also translated worse ($r=\mlRChrfFert$ between tokens per word and chrF++). These findings reveal a double multilingual tax in diffusion decoding: languages already disadvantaged by longer token sequences also require more computation per token, compounding disparities in inference efficiency.

\section{Decoding Cost as a Measure of Certainty}
\label{sec:answer_speed_weighted_vote}

\begin{figure}[t]
\centering
\includegraphics[width=0.9\columnwidth]{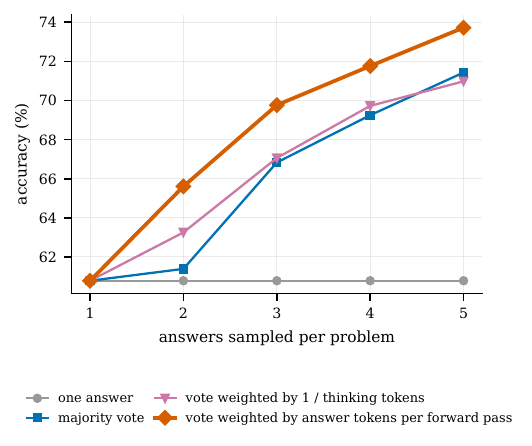}
\caption{Weighting each answer by how fast it was decoded beats the plain majority
vote. (\cvNProblems math problems with \cvNSeeds runs each, DiffusionGemma, thinking on, shipped settings). 
}
\vspace{-0.1in}
\label{fig:answer_speed_weighted_vote_math}
\end{figure}

Section~\ref{sec:harder_problems_and_wrong_answers} showed that a wrong answer takes more forward passes per token than a right one. This gives a free signal for choosing between answers. A common way to raise accuracy is to sample several answers and keep the most frequent one (majority vote; \citealp{wang2023selfconsistency, brown2024largelanguagemonkeysscaling}). We instead weight each answer's vote by how fast the model decoded it.\footnote{Here, the answer is only the text written after the thinking ends; the thinking is not counted.} The weight is the number of answer tokens per forward pass spent on the answer, and costs nothing extra, since the passes are counted while decoding.

We test this on all \cvNProblems competition math problems, with \cvNSeeds runs per problem under the shipped settings (Figure~\ref{fig:answer_speed_weighted_vote_math}). The weighted vote beats the plain majority vote at every
number of answers, by \cvGainMax points with two answers and \cvGainMin with five. The gain is
largest with two answers, where majority vote cannot break a tie, and our weight decides it. With
five answers, accuracy rises from \cvMajFive to \cvWFive; across 10{,}000 bootstrap resamples, the gain remains strictly positive (95\% CI: [\cvGainFiveLo, \cvGainFiveHi] points). Furthermore, this gain is not merely a preference for short reasoning. Weighting by 1 divided by the number of thinking tokens performs similarly to a plain majority vote (\cvThinkFive with five answers). Thus, answer decoding speed serves as a reliable, cost-free measure of model certainty that can directly improve answer selection.


\section{Related Work}

\paragraph{Diffusion LLM interpretability.}
A growing body of work seeks greater understanding and interpretability of diffusion language model decoding.
Work in a similar vein to ours shows that the final answer can be extracted before decoding terminates
\citep{li2025diffusion}, and correct answers appear at
intermediate steps only to be overwritten later \citep{wang2026time}.
Other papers show that contrary to their design, masked diffusion LLMs tend to decode approximately left to right \citep{li2026diffusion}. Uniform diffusion models, such as DiffusionGemma, have been found to exhibit non-chronological reasoning and retroactive self-correction \citep{engels2026transparent}. Analyses of diffusion model latent spaces show that diffusion language models are able to perform induction bidirectionally \citep{catruna2026induction}  and commit to higher-level concepts with different schedules, such as topic before sentiment \citep{zhou2026steering}. While these works explain aspects of diffusion decoding, they do not address the computational question of how diffusion language models allocate their forward passes.

\paragraph{Data-dependent parallelism.}
In our work, we study diffusion language models, which are instances of models capable of parallel decoding. Speculative decoding is another instance of parallel decoding \cite{leviathan2023fast}. In speculative decoding, it is established that parallelism, and therefore speed, depends
on the data. Speculative decoding benchmarks therefore include a variety of tasks to measure speedup in each domain
\citep{li2024eagle,abramovich2026speed, liu2026speculative}. Similar to our work, \citet{sandler2025disparate} analyze the implications of task-dependent speedup in speculative decoding for fairness between different languages. The same data-dependent phenomenon naturally applies to speculative decoding with a diffusion drafter \cite{chen2026dflash}. In the context of self-speculative diffusion, \citet{fu2026nemotron} perform a "speed of light" acceptance rate analysis on different domains and find disparate acceptance rates for each. In pure diffusion decoding, ParallelBench \citep{kang2026parallelbench} argues that masked diffusion adaptive samplers fail to adjust well to task difficulty; however, it does not study more recent uniform diffusion models, such as DiffusionGemma.


\section{Conclusions and Future Work}
\label{sec:conclusions_and_future_work}
We studied where diffusion language models spend their forward passes. Across two models, copying and lightly editing text costs less per token than open-ended writing, with a 3.1-fold gap between grammar correction and story generation in DiffusionGemma. The amount of copies the output has and how common its words are within the task explain part of this difference. The same pattern appears within a response: final answers tend to be cheaper when the thinking has already established what to say. We also find that lower-resource languages pay twice, requiring both more tokens and more passes per token. Finally, wrong math answers cost more to decode than correct ones, even on the same problem. Using this difference to weight votes by final-answer decoding speed improves accuracy by +2.3 percentage points over five-sample majority voting without additional cost.

Future work could use decoding effort to guide verification, additional sampling, and early stopping, or adapt block sizes and stopping rules to task structure and reuse of earlier text. These findings also motivate hybrid models that selectively use autoregressive or diffusion decoding according to what remains to be generated. Taken together, our results show when diffusion can generate efficiently and how its decoding effort can guide inference, opening a path toward models that adapt their decoding strategy and computational budget to the demands of each task and each part of a response.



\newpage

\section*{Limitations}
\paragraph{Few suitable models.}
We study two models because few others fit our question. We need diffusion models that are trained to decode adaptively, deciding at each step how many tokens they are confident enough to commit, and that
are strong enough to handle reasoning, translation, and long-form writing. Such models are very new. DreamReasoner appeared in June 2026 and DiffusionGemma in August 2026. Most earlier open models were
trained with a fixed number of steps, and adaptive decoding was added to them later as an inference trick \citep{fastdllm}. 
As more such models are released, our findings can be extended across wider model scales and training recipes.



\bibliography{custom}

\newpage

\appendix
\counterwithin{figure}{section}
\counterwithin{table}{section}

\section{Models and Settings}
\label{app:models_and_settings}
\begin{table*}[t]
\centering
\small
\begin{tabular}{p{3.5cm}p{5.8cm}p{4.4cm}}
\toprule
 & DiffusionGemma-26B-A4B-it & DreamReasoner-8B \\
\midrule
Weights & \url{google/diffusiongemma-26B-A4B-it} & \url{Dream-org/DreamReasoner-8B} \\
Size & 25.8B total, about 3.8B used per token (mixture of experts) & 8.2B (dense) \\
Tokens written at a time & 256 & 32 \\
Most steps allowed per 256 / 32 tokens & 48 & 32 \\
How it picks tokens to write & lowest entropy first, up to a total entropy of 0.1
  & every token with confidence $\ge 0.9$ \\
\bottomrule
\end{tabular}
\caption{Model settings}
\label{tab:model_settings}
\end{table*}

Both models run with their published decoding settings (Table~\ref{tab:model_settings}). 
We count one forward pass for each denoising step on a block, which is 256 positions for DiffusionGemma and 32 for DreamReasoner. The pass that reads the context is not counted, in the same way that we do not count prompt processing for an autoregressive model. Output tokens are the model's own tokens up to the first \texttt{<eos>}, not counting special tokens. With thinking off, DiffusionGemma's count includes the 2 tokens of its empty thinking header. We compute cost (Eq.~\ref{eq:cost:forward_passes_per_100_tokens}) for each output and average it over the outputs of a task. Error bars are 95\% bootstrap intervals over items with 10{,}000 resamples, and the value for a group of tasks is the average of its task means.

\section{Tasks and Data}
\label{app:tasks_and_data}

Table~\ref{tab:task_data} describes the data. We fixed every item with a selection seed before running any model. To make sure the reproducing tasks really copy from the input, we looked at their gold answers: 96.2\% of the words in a corrected essay and 95.2\% of the words in an extractive summary also appear in the input. To keep output length from varying too much, the extractive summarization prompt asks for as many sentences as the gold summary has (23 at the median), and the story prompt asks for 700--900 words. The three rewriting tasks were run in exactly the same way as the other seven. 

\begin{table*}[t]
\centering
\small
\begin{tabular}{p{2.3cm}p{5.2cm}p{5.4cm}r}
\toprule
task & data & items & $n$ \\
\midrule
\multicolumn{4}{l}{\emph{reproducing}} \\
Grammar fix & W\&I+LOCNESS v2.1, BEA-2019 dev set \citep{bea2019} & whole essays of 80--1{,}200 words & 317 \\
Extractive summ. & GovReport, test set \citep{govreport} & 300 of the 824 reports with 1{,}500--12{,}000 words & 300 \\
\midrule
\multicolumn{4}{l}{\emph{chained}} \\
Code edit & HumanEvalFix, Python \citep{octopack} & the full set & 164 \\
Code gen. & LiveCodeBench \texttt{code\_generation\_lite} v5+v6 \citep{livecodebench} & the 300 newest by contest date & 300 \\
Math & MathArena \citep{matharena}: AIME 2025, AIME 2026, HMMT Feb 2025, HMMT Nov 2025, BRUMO 2025, SMT 2025, CMIMC 2025, APEX 2025 & all problems with a final answer, excluding image and proof problems & 255 \\
\midrule
\multicolumn{4}{l}{\emph{open-ended}} \\
WildBench & WildBench v2 \citep{wildbench} & single-turn creative writing, brainstorming, advice seeking, editing, and information seeking & 294 \\
Story & WritingPrompts, test set \citep{writingprompts} & 300 sampled rows (280 distinct premises) & 300 \\
\midrule
\multicolumn{4}{l}{\emph{rewriting} (from Section~\ref{sec:what_makes_an_output_cheap} on)} \\
Translation & WMT23 general test set, German to English \citep{wmt23}, grouped into documents & all documents of 40 words or more (118 of 210) & 118 \\
Clinical note & ACI-Bench \citep{acibench}: MEDIQA-Chat 2023 Task B test 1, Task C test 2, CLEF Task C test 3 & the full set & 120 \\
Abstractive summ. & BookSum, chapter level, test set \citep{booksum} & 300 of the 669 chapters with 600--8{,}000 words and a 120--900-word summary & 300 \\
\midrule
\multicolumn{3}{l}{total per model} & 2{,}468 \\
\bottomrule
\end{tabular}
\caption{Data for all ten tasks. The first seven appear in Figure~\ref{fig:forward_passes_per_100_tokens_by_task}; the three rewriting tasks are added from Section~\ref{sec:what_makes_an_output_cheap} on.}
\label{tab:task_data}
\label{tab:rewriting_task_data}
\end{table*}




\subsection{Prompts}
\label{app:prompts}

For code editing, code generation and math we use each benchmark's official prompt. WildBench items are given to the model as the user's message, with nothing added. We wrote the remaining prompts ourselves. Text in braces is filled in from the item.

\paragraph{Grammar correction.}
\begin{Verbatim}[fontsize=\scriptsize]
Correct all spelling, grammar, and punctuation errors
in the essay below. Rewrite the entire essay. Change
nothing that is already correct. Do not rephrase for
style, do not add or remove content. Output only the
corrected essay.

Essay:
{text}
\end{Verbatim}

\paragraph{Extractive summarization.}
\begin{Verbatim}[fontsize=\scriptsize]
Below is a government report. Produce an
EXTRACTIVE summary of it.

An extractive summary is built only from sentences
copied out of the source: select exactly
{n_sentences} sentences from the report and
reproduce them verbatim, word for word, in the order
they appear in the report. Do not paraphrase, do not
rewrite, do not merge or split sentences, and do not
write any sentence of your own.

Choose the {n_sentences} sentences that together
best cover the report's findings, recommendations,
and conclusions.

Output only the {n_sentences} selected sentences,
one per line, and nothing else.

Report:
{text}
\end{Verbatim}

\paragraph{Code editing} (HumanEvalFix, official).
\begin{Verbatim}[fontsize=\scriptsize]
{prompt_base}{buggy_solution}
Fix bugs in {entry_point}.

{prompt_base}
\end{Verbatim}

\paragraph{Code generation} (LiveCodeBench, official).
\begin{Verbatim}[fontsize=\scriptsize]
### Question:
{question_content}

### Format: {format_message}
```python
{code_stub}
```

### Answer: (use the provided format with
backticks)
\end{Verbatim}

\paragraph{Math} (MathArena, official). The last line changes with the competition. For AIME it is:
\begin{Verbatim}[fontsize=\scriptsize]
{problem}

Put your final answer within \boxed{}.
The answer is an integer between 0 and 999
inclusive.
\end{Verbatim}

\paragraph{Story.}
\begin{Verbatim}[fontsize=\scriptsize]
Write a complete short story based on the premise
below. Write roughly 700-900 words. Output only the
story.

Premise: {prompt}
\end{Verbatim}

\paragraph{Translation.}
\begin{Verbatim}[fontsize=\scriptsize]
Translate the following German text into English.
Translate the whole text. Output only the English
translation.

German:
{text}
\end{Verbatim}

\paragraph{Clinical note.}
\begin{Verbatim}[fontsize=\scriptsize]
Below is a transcript of a conversation between a
doctor and a patient. Write the clinical note for this
encounter, using the standard section headings (CHIEF
COMPLAINT, HISTORY OF PRESENT ILLNESS, REVIEW OF
SYSTEMS, PHYSICAL EXAM, RESULTS, ASSESSMENT AND PLAN),
each on its own line in capitals. Output only the note.

Transcript:
{text}
\end{Verbatim}

\paragraph{Abstractive summarization.}
\begin{Verbatim}[fontsize=\scriptsize]
Write a summary of the chapter below. Describe what
happens and why it matters to the story. Output only
the summary.

Chapter:
{text}
\end{Verbatim}




\section{Runs and Exclusions}
\label{app:runs_reruns_left_out}

Each model produced one generation per item with seed 1, one item at a time, on NVIDIA H100 GPUs, for \figOneNItems generations in total. When a generation hit the 16{,}384-token limit we ran it again with a limit of 32{,}768 and the same seed, and when one came back empty we ran it again with a new seed. Table~\ref{tab:forward_passes_by_task_counts} gives the counts.

Two kinds of output are left out of the averages. The first are generations that still hit the limit, since their length then reflects the limit and not the task. DiffusionGemma has \dgcodegenCapped of these, all in code generation. DreamReasoner has \dreamstoryCapped stories, \dreamwildbenchCapped WildBench answers, \dreamextsummCapped extractive summaries and \dreammathCapped math solutions. The second are outputs shorter than 5 tokens, which only happened for 2 DreamReasoner stories. Almost all of the long outputs are repetition loops, and loops are cheap to decode. Had we kept them, DreamReasoner's stories would drop from \dreamstoryRho to \dreamstoryRhoWithCapped and look cheaper than its code. For DiffusionGemma it makes little difference, as code generation only moves from \dgcodegenRho to \dgcodegenRhoWithCapped.

\begin{table*}[t]
\centering
\footnotesize
\setlength{\tabcolsep}{3pt}
\begin{tabular}{llrrrrrrrrrr}
\toprule
 & & \multicolumn{4}{c}{items} & \multicolumn{3}{c}{cost} & \multicolumn{2}{c}{output tokens} & \\
\cmidrule(lr){3-6}\cmidrule(lr){7-9}\cmidrule(lr){10-11}
model & task & total & used & hit limit & empty & mean & 95\% CI & median & mean & median & tokens/pass \\
\midrule
DiffusionGemma & Grammar fix & 317 & 317 & 0 & 0 & 3.72 & [3.58, 3.86] & 3.49 & 282 & 237 & 28.8 \\
 & Extractive summ. & 300 & 300 & 0 & 0 & 4.34 & [4.21, 4.47] & 4.18 & 874 & 798 & 24.4 \\
 & Code edit & 164 & 164 & 0 & 0 & 5.03 & [4.76, 5.33] & 4.73 & 405 & 416 & 20.3 \\
 & Math & 255 & 255 & 0 & 0 & 6.18 & [5.94, 6.41] & 6.00 & 1{,}340 & 1{,}130 & 15.0 \\
 & Code gen. & 300 & 295 & 5 & 0 & 6.21 & [5.96, 6.45] & 5.75 & 1{,}733 & 675 & 12.1 \\
 & WildBench & 294 & 294 & 0 & 0 & 8.25 & [7.96, 8.57] & 7.66 & 874 & 847 & 12.4 \\
 & Story & 300 & 300 & 0 & 0 & 11.43 & [11.30, 11.55] & 11.42 & 1{,}041 & 1{,}058 & 8.7 \\
\midrule
DreamReasoner & Extractive summ. & 300 & 260 & 40 & 0 & 22.05 & [20.85, 23.44] & 19.76 & 4{,}802 & 4{,}716 & 4.7 \\
 & Grammar fix & 317 & 317 & 0 & 0 & 22.21 & [21.41, 23.05] & 20.76 & 288 & 242 & 4.7 \\
 & Math & 255 & 250 & 5 & 0 & 52.24 & [50.76, 53.72] & 50.74 & 5{,}050 & 3{,}185 & 1.9 \\
 & Code gen. & 300 & 300 & 0 & 0 & 56.50 & [55.15, 57.79] & 59.78 & 2{,}088 & 1{,}778 & 1.6 \\
 & Code edit & 164 & 164 & 0 & 0 & 59.94 & [58.65, 61.21] & 60.38 & 783 & 780 & 1.7 \\
 & WildBench & 294 & 238 & 56 & 0 & 78.57 & [75.45, 81.63] & 85.19 & 2{,}118 & 1{,}522 & 1.6 \\
 & Story & 300 & 155 & 143 & 2 & 95.47 & [93.69, 96.99] & 98.50 & 1{,}605 & 1{,}420 & 1.1 \\
\bottomrule
\end{tabular}
\caption{The numbers behind Figure~\ref{fig:forward_passes_per_100_tokens_by_task}. Cost is forward passes per 100 output tokens (Eq.~\ref{eq:cost:forward_passes_per_100_tokens}), averaged over the items we used. We leave out items that were still writing when they reached the token limit, and items with fewer than 5 output tokens, where a per-token cost is not meaningful. The last column divides all output tokens by all forward passes over the items used. Within each model, rows are sorted by mean cost.}
\label{tab:forward_passes_by_task_counts}
\end{table*}

\section{Multilingual Runs}
\label{app:multilingual_runs}

We gave DiffusionGemma the first 200 English sentences of FLORES-200 \citep{nllb2022} and asked it to translate each one into \mlNLang languages written in 16 scripts, with thinking off, seed 1 and a 4{,}096-token limit. This gave \mlNRecords generations, all of which finished, so nothing was left out. Fertility (tokens per word) and length relative to English are measured on the reference translations with the model's tokenizer, which means they are fixed before the model runs. We do not compute fertility for the five languages written without spaces between words (Chinese, Japanese, Thai, Khmer and Burmese). Translation quality is measured with chrF++ against the reference.

\begin{figure}[t]
\centering
\includegraphics[width=\columnwidth]{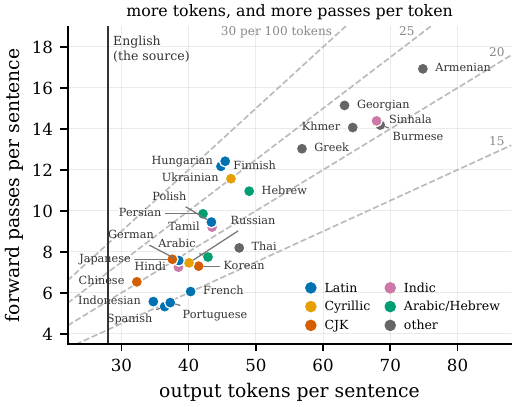}
\caption{\textbf{Some languages cost more twice: more tokens, and more passes per token.} DiffusionGemma translates the same 200 English FLORES sentences into \mlNLang languages. Each point is one language, coloured by script, with tokens per sentence on the x-axis and forward passes per sentence on the y-axis. Dashed lines mark constant cost, so a point on a higher line needs more passes per token. The vertical line shows English for reference (\mlEnTokens tokens per sentence); it is the source language, so it has no cost. All values are in Table~\ref{tab:multilingual_cost_by_language}.}
\label{fig:multilingual_tokens_vs_passes_by_language}
\end{figure}

\begin{table*}[t]
\centering
\footnotesize
\setlength{\tabcolsep}{2.5pt}
\begin{tabular}{l rrr rrr}
\toprule
language & tokens & passes & cost & fertility & vs.\ En & chrF++ \\
\midrule
Spanish & 36.4 & 5.33 & 15.7 & 1.71 & 1.29 & 54.5 \\
Portuguese & 37.3 & 5.52 & 15.9 & 1.82 & 1.26 & 69.0 \\
Indonesian & 34.7 & 5.57 & 17.4 & 2.03 & 1.16 & 68.4 \\
French & 40.3 & 6.06 & 16.3 & 1.75 & 1.41 & 68.2 \\
Chinese & 32.3 & 6.54 & 21.9 & -- & 1.12 & 32.0 \\
Hindi & 38.5 & 7.25 & 20.2 & 1.79 & 1.31 & 54.0 \\
Korean & 41.5 & 7.30 & 19.1 & 2.72 & 1.39 & 36.2 \\
Russian & 40.0 & 7.46 & 20.5 & 2.47 & 1.40 & 56.1 \\
German & 38.5 & 7.58 & 21.0 & 2.04 & 1.34 & 63.7 \\
Japanese & 37.6 & 7.63 & 21.7 & -- & 1.24 & 35.9 \\
Arabic & 42.9 & 7.75 & 19.6 & 2.46 & 1.48 & 55.3 \\
Thai & 47.5 & 8.19 & 19.0 & -- & 1.65 & 46.6 \\
Tamil & 43.5 & 9.21 & 22.9 & 3.08 & 1.49 & 45.3 \\
Polish & 43.4 & 9.45 & 22.9 & 2.53 & 1.52 & 50.0 \\
Persian & 42.1 & 9.86 & 25.0 & 2.09 & 1.45 & 47.7 \\
Hebrew & 49.0 & 10.96 & 24.0 & 2.97 & 1.73 & 47.8 \\
Ukrainian & 46.3 & 11.57 & 26.9 & 2.69 & 1.64 & 52.6 \\
Finnish & 44.8 & 12.17 & 28.8 & 3.11 & 1.67 & 49.8 \\
Hungarian & 45.4 & 12.42 & 29.5 & 2.79 & 1.70 & 42.0 \\
Greek & 56.9 & 13.03 & 24.3 & 2.82 & 2.22 & 41.3 \\
Khmer & 64.4 & 14.06 & 23.5 & -- & 2.85 & 32.1 \\
Burmese & 68.5 & 14.19 & 22.6 & -- & 2.42 & 36.8 \\
Sinhala & 68.0 & 14.39 & 22.9 & 3.46 & 2.35 & 36.5 \\
Georgian & 63.2 & 15.14 & 25.9 & 3.95 & 2.32 & 37.4 \\
Armenian & 74.8 & 16.93 & 24.7 & 4.11 & 2.89 & 33.0 \\
\bottomrule
\end{tabular}
\caption{Results for each language, averaged over the same 200 FLORES sentences and sorted by forward passes per sentence. \emph{Tokens} and \emph{passes} are per sentence, and cost is forward passes per 100 tokens. Fertility (tokens per word) and length relative to English are measured on the reference translations; fertility is not defined for languages written without spaces.}
\label{tab:multilingual_cost_by_language}
\end{table*}

\section{An Example of Self-Correction}
\label{app:self_correction_example}

Figure~\ref{fig:self_correction_example_right_vs_wrong} shows two answers to the same competition math problem that differ only in the random seed.

\section{E1 elaboration: Use of AI Assistants}
We used AI assistants to polish manuscript text, find relevant citations and related work, and assist with code generation and plotting.

\begin{figure*}[t]
\centering
{\sethlcolor{selfhl}
\fbox{\begin{minipage}[t]{0.485\textwidth}\footnotesize\raggedright\textbf{Seed 1: correct (4529), 3.8 passes per 100 tokens}\par\smallskip To compute the sum $A = \sum_{a=1}^{96} \sum_{n=1}^{96} \left\{\frac{a^{n}}{97}\right\}$, we note that $\left\{\frac{a^n}{97}\right\} = \frac{a^n \pmod{97}}{97}$. Thus, $\hookleftarrow$ \ldots{} $\hookleftarrow$ Then, $A = \frac{S}{97} = \frac{96}{97} + 4528 = 4528.9896...$ $\hookleftarrow$ Rounding to the nearest integer, we get 4529. $\hookleftarrow$ $\boxed{4529}$\end{minipage}}\hfill\fbox{\begin{minipage}[t]{0.485\textwidth}\footnotesize\raggedright\textbf{Seed 2: wrong (4584), 7.0 passes per 100 tokens}\par\smallskip To compute the sum $A = \sum_{a=1}^{96} \sum_{n=1}^{96} \left\{\frac{a^{n}}{97}\right\}$, we let $p=97$. The fractional part $\left\{\frac{a^n}{p}\right\}$ is equal to $\frac{a^n \pmod{p}}{p}$. Thus, $\hookleftarrow$ \ldots{} $\hookleftarrow$ \hl{Wait}, the previous definition was $V_d$ as the set of $(p-1)/d$-th powers. Let's re-evaluate: \ldots{} $\hookleftarrow$ For $p=97$, $A \approx 4608 - 48 = 4560$. $\hookleftarrow$ \hl{Actually}, the sum $S = \sum_{a=1}^{p-1} \sum_{n=1}^{p-1} (a^n \pmod{p})$ is known to be $\frac{p(p-1)^2}{2} - \frac{p(p-1)}{4}$ in certain cases. \ldots{} $\hookleftarrow$ For $p=97$, $A \approx 4608 - 24 = 4584$. $\hookleftarrow$ $\boxed{4584}$\end{minipage}}}
\caption{\textbf{One problem, answered correctly with one seed and incorrectly with another.} DiffusionGemma with thinking on, on \texttt{brumo\_2025-15}. We show the start and end of each final answer and the passages around its first self-correction phrases (highlighted); \ldots{} marks omitted text and $\hookleftarrow$ a line break. The correct answer restates what the thinking found. The wrong one keeps working on the problem as it writes, and costs almost twice as many passes per token.}
\label{fig:self_correction_example_right_vs_wrong}
\end{figure*}

\end{document}